\def\year{2017}\relax
\documentclass[letterpaper]{article}
\usepackage{aaai17}
\usepackage{times}
\usepackage{helvet}
\usepackage{courier}

\usepackage{epsfig}
\usepackage{booktabs}
\usepackage{colortbl}
\usepackage{multirow}
\usepackage{natbib}
\usepackage{amsmath}
\usepackage{algorithm}
\usepackage{algpseudocode}
\newcommand{\myrowcolour}{\rowcolor[gray]{0.925}}

\begin{document}
%
\title{Sparse Boltzmann Machines \\with Structure Learning as Applied to Text Analysis}
\author{Zhourong Chen\thanks{Corresponding authors.} \quad Nevin L. Zhang\footnotemark[1] \quad Dit-Yan Yeung \quad Peixian Chen \\
Hong Kong University of Science and Technology \\
\{zchenbb,lzhang,dyyeung,pchenac\}@cse.ust.hk
}
\maketitle
\begin{abstract}
We are interested in exploring the possibility and benefits of structure learning for deep models. As the first step, this paper investigates the matter for \emph{Restricted Boltzmann Machines (RBMs)}. We conduct the study with Replicated Softmax, a variant of RBMs for unsupervised text analysis. We present a method for learning what we call \emph{Sparse Boltzmann Machines}, where each hidden unit is connected to a subset of the visible units instead of all of them.  Empirical results show that the method yields models with significantly improved model fit and interpretability as compared with RBMs where each hidden unit is connected to all visible units.
\end{abstract}

\section{Introduction}
Deep learning has achieved great successes in recent years. It has produced superior results in a range of applications, including image classification \citep{krizhevsky2012imagenet}, speech recognition \citep{hinton2012deepsppech,mikolov2011strategies}, language translation \citep{sutskever2014sequence} and so on. It is now time to ask whether it is possible and beneficial to learn structures for deep models.

To learn the structure of a deep model, we need to determine the number of hidden layers and the number of hidden units at each layer. More importantly, we need to determine the connections between neighboring layers. This implies that we need to talk about sparse models where neighboring layers are not fully connected.

Sparseness is desirable and full connectivity is unnecessary. In fact, \cite{NIPS2015_5784} have shown that many weak connections in the fully connected layers of \emph{Convolutional Neural Networks (CNNs)} \citep{Lecun98gradient-basedlearning} can be pruned without incurring any accuracy loss.  The convolutional layers of CNNs are sparse, and the fact is considered one of the key factors that lead to the success of CNNs.  Moreover, it is well known that overfitting is a serious problem in deep models. One method to address the problem is dropout \citep{Srivastava:2014:DSW:2627435.2670313}, which randomly drops out units (while keeping full connectivity) during training. The possibility of randomly dropping connections has also been explored in \cite{wan2013regularization}. Sparseness offers an interesting alternative. It amounts to deterministically drop out connections.

How can one learn sparse deep models? One method is to first learn a fully connected model and then prune weak connections \citep{NIPS2015_5784}. The drawbacks of this method are that it is computationally wasteful and doesn't provide a way to determine the number of hidden units. We would like to develop a method that determines the number of hidden units and the connections between units automatically. The key intuition is that a hidden unit should be connected to a group of strongly correlated units at the level below. This idea is used in convolutional layers of CNNs, where a unit is connected to pixels in a small patch of an image. In image analysis, spatial proximity implies strong correlation.

To apply the intuition to applications other than image analysis, we need to identify groups of strongly correlated variables for which latent variables should be introduced. \emph{Hierarchical Latent Tree Analysis (HLTA)} (Liu et al 2014, Chen et al 2016) offers a plausible solution. HLTA first partitions all the variables into groups such that the variables in each group are strongly correlated and the correlations can be properly modelled using a single latent variable. It introduces a latent variable for each group. Then it converts the latent variables into observed variables via data completion and repeats the process to produce a hierarchy. The output of HLTA is a hierarchical latent tree model where the observed variables are at the bottom and there are multiple layers of latent variables on top. To obtain a non-tree sparse deep model, we propose to use the tree model as a skeleton and introduce additional connections to model the residual correlations not captured by the tree.

In this paper, we fully develop and test the idea in the context of RBMs, which have a single layer of hidden units and are building blocks of Deep Belief Networks and Deep Boltzmann Machines.  The target domain is unsupervised text analysis. We present an algorithm for learning what we call Sparse Boltzmann Machines. Empirically, we show that the full-connectivity restriction of RBMs can easily lead to overfitting, and that Sparse Boltzmann Machines are effective in avoiding overfitting. We also demonstrate that Sparse Boltzmann Machines are more interpretable than RBMs.

\section{Related Works}
The concept of sparse RBMs were first mentioned in \citep{lee2008sparse}. The authors use sparse RBMs to build sparse Deep Belief Networks and extract some interesting features. However, in their paper, sparse RBMs were not defined from the perspective of sparse connections but sparse hidden unit activations. And it was achieved by adding a regularization term to the objective function when training the parameters. There is no structure learning.

Network pruning is also a potential way to optimize the structure of a neural network. Biased weight decay was the early approach to pruning. Later, Optimal Brain Damage \citep{Cun90optimalbrain} and Optimal Brain Surgeon \citep{Hassibi93secondorder} suggested that magnitude-based pruning may not be the best strategies and they proposed pruning methods based on the Hessian of the loss function. With respect to deep neural networks, \cite{NIPS2015_5784} proposed to compress a network through a three-step process: train, prune connections, and retrain. We call it redundancy pruning. In contrast, \cite{Srinivas2015} proposed to prune redundant neurons directly. They all reduced the number of parameters vastly with slight or even no performance loss. The drawback of network pruning is that the original networks should be large enough and hence some computation would be wasted on those unnecessary parameters during pre-training.

\section{Restricted Boltzmann Machines}

\begin{figure}[t]
\begin{center}
\includegraphics[width=4cm]{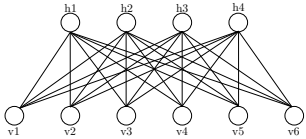}
\vspace{-0.3cm}
\caption{An example RBM with $K=6$ and $F=4$.}
\vspace{-0.5cm}
\label{fig.rbm}
\end{center}
\end{figure}

An \emph{Restricted Boltzmann Machine (RBM)} \citep{Smolensky:1986:IPD:104279.104290} is a two-layer undirected graphical model with a layer of $K$ visible units $\{v^1, \ldots, v^K\}$ and a layer of $F$ hidden units $\{h_1, \ldots, h_F\}$. The two layers are fully connected to each other, while there are no connections between units at the same layer. An example is shown in Figure \ref{fig.rbm}. In the simplest case, all the units are assumed to be binary. An energy function is defined over all the units as follows:

\vspace{-0.55cm}
\begin{small}
\begin{equation}
E({\bf v}, {\bf h}) = - \sum_{j=1}^F\sum_{k=1}^K  W^k_jh_jv^k  - \sum_{k=1}^K  v^kb^k - \sum_{j=1}^F  h_ja_j
\end{equation}
\end{small}
\vspace{-0.25cm}

\noindent where $a_j$ and $b^k$ are bias parameters for the hidden and visible units respectively, while $W^k_j$ is the connection weight between hidden unit $h_j$ and visible unit $v^k$. The energy function defines a joint probability over \textbf{v} and \textbf{h} as follows: \begin{equation}P(\textbf{v},\textbf{h}) = exp(-E(\textbf{v},\textbf{h}))/ Z\end{equation}
where \begin{small}$Z=\sum_{\textbf{v}',\textbf{h}'}exp(-E(\textbf{v}',\textbf{h}'))$\end{small} is a normalization term called the partition function.  An important property of RBM is that the conditional distributions $P(\textbf{h}|\textbf{v})$ and $P(\textbf{v}|\textbf{h})$ factorize as below:
\begin{small}
\begin{equation}P(\textbf{h}|\textbf{v}) = \prod_{j}P(h_{j}|\textbf{v}) \hspace{20pt} P(\textbf{v}|\textbf{h}) = \prod_{k}P(v^{k}|\textbf{h})\end{equation}
\vspace{-0.3cm}
\begin{equation}P(h_j=1|{\bf v})  = \sigma(a_j +\sum_{k=1}^K W^k_jv^k ) \end{equation}
\begin{equation}P(v^k=1|{\bf h})  = \sigma(b^k +\sum_{j=1}^F W^k_jh_j ) \end{equation}
\end{small}

\noindent where $\sigma(x) = 1/(1+e^{-x})$ is the logistic function. The model parameters of an RBM are learned using the Contrastive Divergence algorithm \citep{Hinton:02}, which maximizes the data likelihood via stochastic gradient descent.

In \cite{NIPS2009_3856}, RBM was used for topic modeling and the proposed model was called Replicated Softmax. Suppose the vocabulary size is $K$. Let us represent a document with $D$ tokens as a binary matrix $\cal{U}$ of size $K * D$ with $u_i^k=1$ if the $i^{th}$ token is the $k^{th}$ word in the vocabulary. The energy function of document $\cal{U}$ and hidden units $\mathbf{h}$ is defined as follows:

\vspace{-0.3cm}
\begin{small}
\begin{eqnarray}
E({\cal U}, {\bf h}) =  - \sum_{j=1}^F\sum_{k=1}^K  W^k_jh_j\hat{u}^k  - \sum_{k=1}^K  \hat{u}^kb^k - D\sum_{j=1}^F  h_ja_j
\end{eqnarray}
\end{small}
\vspace{-0.1cm}

\noindent where $\hat{u}^k = \sum_{i=1}^D u_i^k$ denotes the count for the $k^{th}$ word. The conditional probabilities $P(h_j=1|{\cal U})$ can be calculated as:
\vspace{-0.3cm}
\begin{small}
\begin{eqnarray}
P(h_j=1|{\cal U}) = \sigma(Da_j + \sum_{k=1}^{K}W_j^k\hat{u}^k)  \label{equ.conditional}
\end{eqnarray}
\end{small}
\vspace{-0.35cm}

\noindent The motivation behind Replicated Softmax is to properly model word counts in documents of varying lengths through weight sharing. It was shown to generalize better than \emph{Latent Dirichlet Allocation (LDA)} \citep{Blei:2003:LDA:944919.944937} in terms of log-probability on held-out documents and accuracy on retrieval tasks. In this paper, we will use Replicated Softmax for text analysis.
\begin{figure}[t]
\begin{center}
\begin{tabular}{c}
\mbox{
        \epsfxsize=3.6cm
        \epsffile{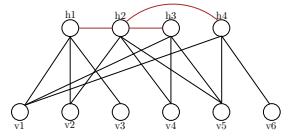}
}
\end{tabular}
\vspace{-0.2cm}
\caption{An example SBM with $K=6$ and $F=4$.}
\vspace{-0.95cm}
\label{fig.sbm}
\end{center}
\end{figure}

\begin{figure*}[t]
\begin{center}
\begin{tabular}{c}
\mbox{
        \epsfxsize=4cm
        \epsffile{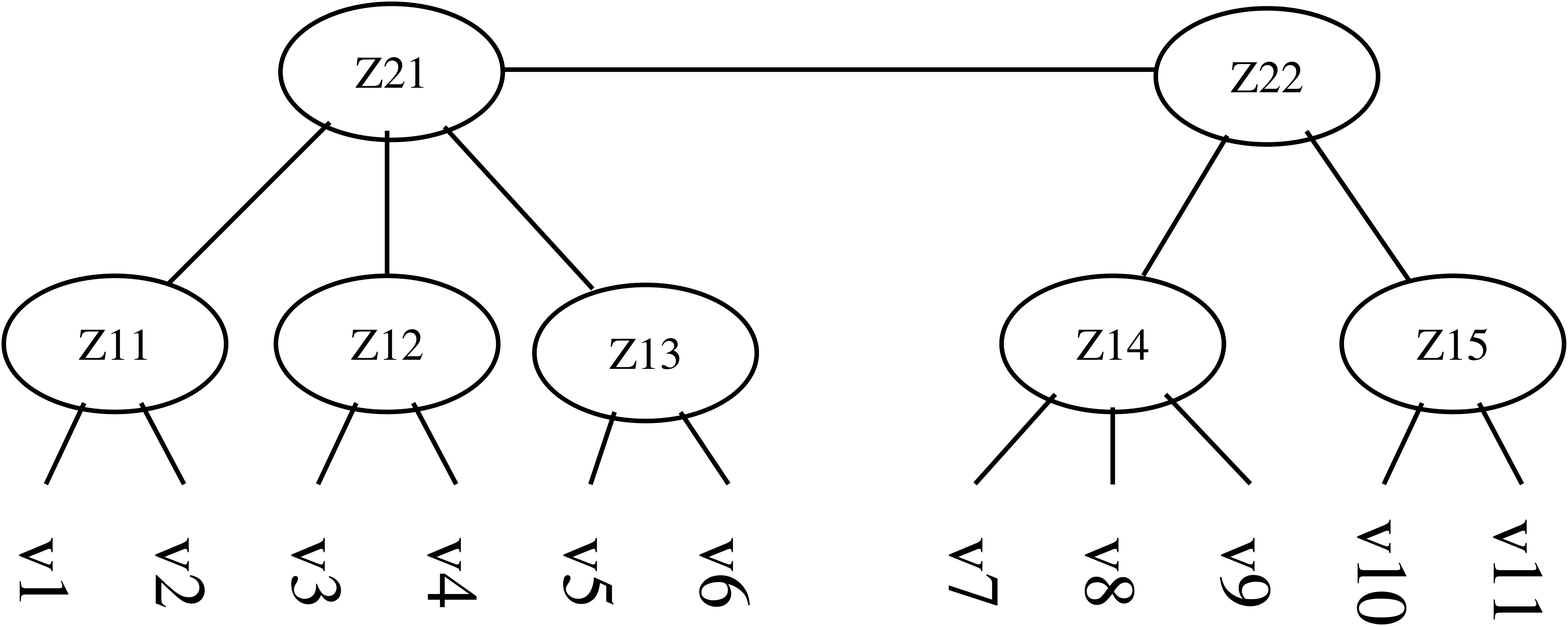}

        \hspace{0.5cm}
        \epsfxsize=6cm
        \epsffile{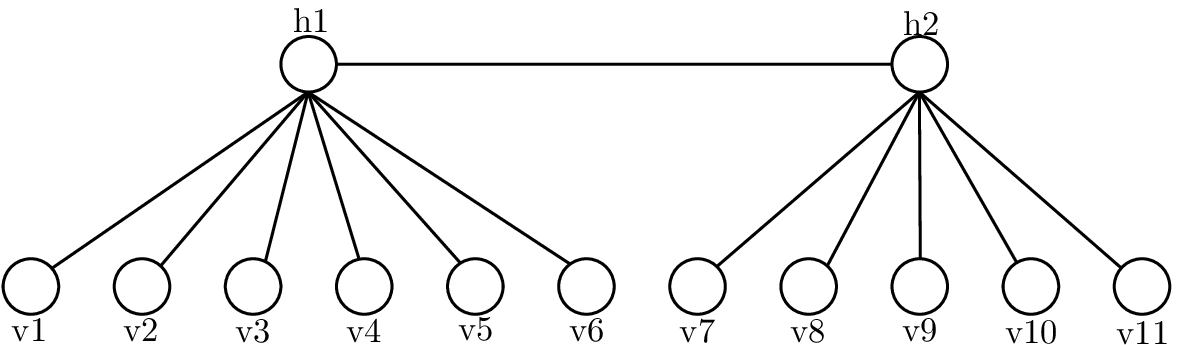}

        \hspace{0.5cm}
        \epsfxsize=6cm
        \epsffile{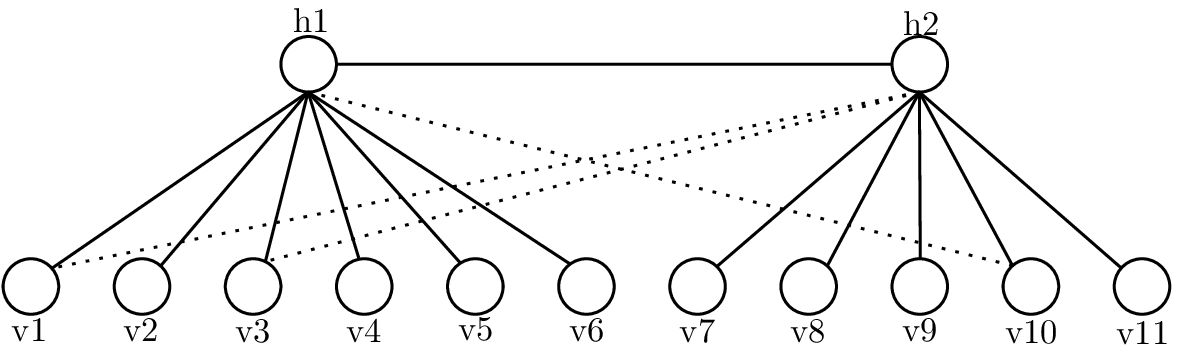}
}
\end{tabular}
\vspace{-0.3cm}
\caption{Structure learning for SBMs: A three layer HLTM is first learned (left).  The hidden variables at the top level are used to build a skeleton for an SBM (middle). An SBM is finally obtained by adding connections to the skeleton (right).}
\label{fig.structure_process}
\vspace{-0.5cm}
\end{center}
\end{figure*}

\section{Sparse Boltzmann Machines}

In this section, we will propose our new models, \emph{Sparse Boltzmann Machines (SBMs)}. An SBM is a two-layer undirected graphical model with a layer of $K$ visible units $\{v^1, \ldots, v^K\}$ and a layer of $F$ hidden units $\{h_1, \ldots, h_F\}$. The hidden units in SBMs are directly linked up to form a tree structure, while each hidden unit is also individually connected to a subset of the visible units. See Figure \ref{fig.sbm} for an example SBM. In SBM, the number of hidden units and the connectivities are both learned from data.

One technical difference between SBMs and RBMs is that there are direct connections among the hidden units in SBMs. We call them hidden connections. The reason why we introduce the hidden connections into our models is that, the hidden connections provide a way to relate a hidden unit to a visible unit without a direct connection. For example, in Figure \ref{fig.sbm}, hidden unit $h_1$ is not directly connected to visible unit $v_4$. However, the existence of the hidden connection between $h_1$ and $h_2$ introduces a path connecting $h_1$ and $v_4$, which can help us to better model the correlation between the two units. This is crucial in reducing the number of connections between hidden units and visible units. To avoid the connections among the hidden units becoming too dense, we restrict them to form a tree structure.


\begin{figure}[t]
\begin{center}
\hspace{-0.4cm}
\includegraphics[width=8.5cm]{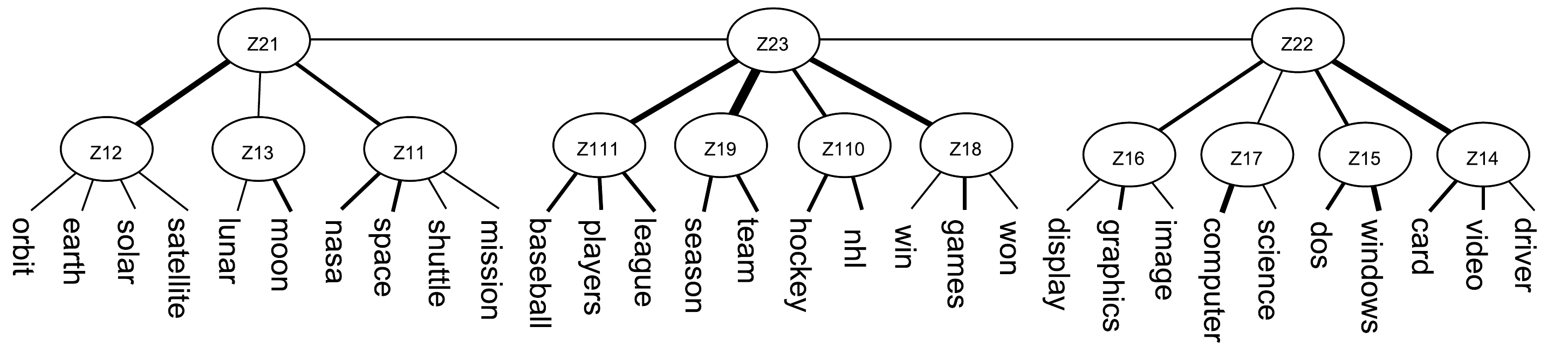}
\vspace{-0.4cm}
\caption{An example HLTM from \cite{DBLP:conf/aaai/ChenZPC16}.}
\vspace{-0.7cm}
\label{fig.HLTM}
\end{center}
\end{figure}

\subsection{Parameter Learning}
SBMs also can be extended for text analysis as RBMs are extended to Replicated Softmax. Here we will introduce SBMs in the context of Replicated Softmax and use the same notations in the previous section. Let $\cal G$ be a graph representing the model structure. Edge $(j, k)$ belongs to $\cal G$ if and only if there is a link between the visible unit $v^k$ and hidden unit $h_j$. Edge $(j, l)$ belongs to $\cal G$ if and only if there is a link between the hidden unit $h_j$ and hidden unit $h_l$. Also let $W_{jl}$ be the weight on the connection between hidden unit $h_j$ and hidden unit $h_l$. Then the energy function of an SBM for document $\cal U$ and hidden units $\mathbf{h}$ is as below:
\begin{small}
\begin{eqnarray}
\begin{aligned}
E({\cal U}, {\bf h}) = & - {\sum_{(j,k) \in \cal G}}W^k_jh_j\hat{u}^k - \sum_{k=1}^K\hat{u}^kb^k \\ & - D\sum_{j=1}^F  h_ja_j - D{\sum_{(j,l) \in \cal G}}W_{jl}h_jh_l.
\end{aligned}
\end{eqnarray}
\end{small}

\noindent Similar to Replicated Softmax, our model defines the joint distribution as:
\begin{small}
\begin{eqnarray}
P({\cal U}, {\bf h}) = \frac{1}{Z} \exp(-E({\cal U}, {\bf h})), \mbox{   }
\end{eqnarray}
\end{small}

\noindent where \begin{small}$Z=\sum_{\cal U'}\sum_{\bf h}  \exp(-E({\cal U'}, {\bf h}))$\end{small}. Note that the summation over $\cal U'$ is done over all the possible documents with the same length as $\cal U$. 

Let $\bar{\cal U} =\{{\cal U}_n\}_{n=1}^N$ be a collection of $N$ documents with potentially different lengths $D_1$, \ldots, $D_N$. We assume that $P(\bar{\cal U}) = \prod_{n=1}^N P({\cal U}_n)$, where $P({\cal U}_n) = \sum_{\bf h}P({\cal U}_n, \bf h)$. The objective of training an SBM for $\bar{\cal U}$ is to maximize the log-likelihood of the documents $\log P(\bar{\cal U})$. We maximize the objective function via stochastic gradient descent. The partial derivatives of $\log P(\bar{\cal U})$ w.r.t the parameters $W_j^k$, $b^k$ and $a_j$ remain the same as in Replicated Softmax:
\begin{small}
\begin{eqnarray}
\begin{aligned}
\frac{\partial \log P (\bar{\cal U})}{\partial W_j^k} = \sum_{n=1}^N (E_{P(h_j|{\cal U}_n)}[h_j\hat{u}^k_{n}] - E_{P({\cal U}, {\bf h})}[h_j\hat{u}^k])
\end{aligned}
\end{eqnarray}

\begin{eqnarray*}
\frac{\partial \log P (\bar{\cal U})}{\partial b^k}
&=&  \sum_{n=1}^N  (\hat{u}^k_n - E_{P({\cal U})}[\hat{u}^k]) \\
\frac{\partial \log P (\bar{\cal U})}{\partial a_j}
&=&  \sum_{n=1}^N  D_{n}(E_{P(h_j|{\cal U}_{n})}[h_j] - E_{P(h_j)}[h_j])
\end{eqnarray*}

\end{small}

\noindent while the partial derivative of $\log P(\bar{\cal U})$ w.r.t the new parameter $W_{jl}$ for fixed $j$ and $l$ is: 
\begin{small}
\begin{eqnarray}
\frac{\partial \log P(\bar{\cal U})}{\partial W_{jl}} = \sum_{n=1}^N  D_{n}(E_{P(\mathbf h|{\cal U}_{n})}[h_jh_l] - E_{P({\mathbf h})}[h_jh_l])
\end{eqnarray}
\end{small}

The first terms in these partial derivatives require the computation of the conditional probabilities $P(h_j|{\cal U}_{n})$ and $P(\mathbf{h}|{\cal U}_{n})$. In Replicated Softmax, $P(h_j|{\cal U}_{n})$ can be calculated using Equation (\ref{equ.conditional}). While in SBMs, due to the connections between hidden units, $P(\mathbf{h}|{\cal U}_{n})$ no longer factorizes and hence Equation (\ref{equ.conditional}) cannot be applied. Nevertheless, since the hidden units in Sparse Boltzmann Machines are linked as a tree structure, we can easily compute the value of $P(h_j|{\cal U}_{n})$ and $P(\mathbf{h}|{\cal U}_{n})$ by conducting message propagation \citep{38136} in the model.

The second terms in these derivatives require taking an expectation with respect to the distribution defined by the model, which is intractable. Thus as in Replicated Softmax, we adopt the Contrastive Divergence algorithm to approximate the second terms by running Gibbs sampling chains in the model. Specifically, the Gibbs chains are initialized at the training data and run for $T$ full steps to draw samples from the model. In SBMs, given a document $\cal U$ and the value of all the other hidden units $\mathbf{h}_{-j}$, the conditional probability to sample a hidden unit $h_j$ becomes:
\begin{small}
\vspace{-0.1cm}
\begin{eqnarray*}
\begin{aligned}
P(h_j=1|{\cal U}, \mathbf{h}_{-j}) = \sigma(& \sum_{(j,k) \in \cal G}  W^k_j\hat{u}^k + Da_j +\\ 
											& D \sum_{(j,l) \in \cal G}W_{jl}h_l +D \sum_{(l,j) \in \cal G}W_{lj}h_l )
\end{aligned}
\end{eqnarray*}
\end{small}

\noindent while the conditional probability to sample an visible unit remains the same as in Replicated Softmax.

\subsection{Structure Learning}
We regard SBMs as a method to model correlations among the visible units. Learning an SBM hence amounts to building a latent structure to explain the correlations. Recently, \cite{DBLP:conf/pkdd/LiuZC14} and \cite{DBLP:conf/aaai/ChenZPC16} proposed a method, called HLTA, for learning a \emph{Hierarchical Latent Tree Model (HLTM)} from data. Our structure learning algorithm for SBMs is built upon their work. We expand the tree model from HLTA to obtain the structure of an SBM.

HLTA learns a tree model ${\cal T}$ with a layer of observed variables at the bottom and multiple layers of latent variables. Note that the visible units and hidden units in SBMs are called observed variables and latent variables in HLTM respectively. The left panel in Figure \ref{fig.structure_process}  and Figure \ref{fig.HLTM} illustrate example models that HLTA produces. Each latent variable in the model is connected to a set of highly-correlated variables in the layer below. The number of latent variables at each layer is determined automatically by the algorithm. The number $L$ of latent layers in ${\cal T}$ is controllable. In this paper, we set $L=2$. Let $H_{l}$ be the $l^{th}$ latent layer in ${\cal T}$. Also let $\mathbf{V}_Z$ be the set of observed variables which are located in the subtree rooted at latent variable $Z$ in ${\cal T}$.

To build the structure of an SBM from ${\cal T}$, we first remove all the latent layers except the top layer $H_{L}$. Then we connect each latent variable $Z$ in $H_{L}$ to the set of observed variables $\mathbf{V}_Z$. We use the resulting structure as a skeleton ${\cal T'}$ of the corresponding SBM. This is illustrated in Figure \ref{fig.structure_process}, where the hidden units $h_1$, $h_2$ in SBM correspond to $Z_{21}$, $Z_{22}$ in ${\cal T}$ respectively. Note that the skeleton is still a tree structure, where each node has only one parent.

As to remove the tree-structure constraint, we conduct an expansion step to increase the number of ``fan-out'' connections for each hidden unit in ${\cal T'}$. The key question is how to determine the new set of visible units that a hidden unit should be connected to. We introduce our method using $Z_{21}$ (correspondingly $h_1$ in ${\cal T'}$) and $v_7$ in Figure \ref{fig.structure_process} as an example. To determine whether $Z_{21}$ should also be connected to $v_7$, we consider the empirical conditional mutual information $I(Z_{21}, v_7|Z_{22},\bar{\cal U})$, where $Z_{22}$ is the root of the subtree that $v_7$ is in. To estimate the value, we first estimate the empirical joint distribution $\hat{p}(Z_{21}, Z_{22}, v_7)$. We go through all the documents and compute $p(Z_{21},Z_{22}|{\cal U}_{n})$ for each document ${\cal U}_{n}$ in $\bar{\cal U}$ by conducting inference in ${\cal T}$. Then we collect the statistics of $Z_{21},Z_{22}$ and $v_7$ to get $\hat{p}(Z_{21}, Z_{22}, v_7)$. After that, $I(Z_{21}, v_7|Z_{22},\bar{\cal U})$ can be estimated as:
\begin{small}
\vspace{-0.2cm}
\begin{eqnarray*}
\begin{aligned}
&I(Z_{21}, v_7|Z_{22},\bar{\cal U}) = \\
&\sum_{Z_{22}}\hat{p}(Z_{22})\sum_{v_7}\sum_{Z_{21}}\hat{p}(Z_{21}, v_7|Z_{22})log \frac{\hat{p}(Z_{21}, v_7|Z_{22})}{\hat{p}(Z_{21}|Z_{22})\hat{p}(v_7|Z_{22})}.
\end{aligned}
\end{eqnarray*}
\end{small}

\noindent All the distributions in the above formula can be derived from the joint distribution $\hat{p}(Z_{21}, Z_{22}, v_7)$.

If the correlation between $Z_{21}$ and $v_7$ is properly modeled in ${\cal T}$, the two variables should be conditionally independent given $Z_{22}$, and hence $I(Z_{21}, v_7|Z_{22},\bar{\cal U})$ should be zero. Therefore, if $I(Z_{21}, v_7|Z_{22}, \bar{\cal U})$ is not 0, then we can conclude that the correlation between $Z_{21}$ and $v_7$ is not properly modeled in the model, and the model needs to be expanded by adding new connections between the two variables.

Our algorithm, called \emph{SBM-SFC (SBM-Structure from Correlation)}, is given in Algorithm 1. It considers the latent variables one at a time. For a given latent variable $Z$ (suppose the corresponding hidden unit in ${\cal T'}$ is $h$), it computes the conditional mutual information between $Z$ and each unconnected observed variable, and sorts the observed variables in descending order with respect to the conditional mutual information. Then in ${\cal T'}$, it connects hidden unit $h$ to the visible units corresponding to the top $M$ observed variables with the highest conditional mutual information. $M$ is a predefined parameter, which normally is set to the value such that each hidden unit is connected to $0.2*K$ hidden units. After the above expansion step is done for each hidden unit in ${\cal T'}$, the whole structure of an SBM is determined.

\begin{algorithm}[t]
\vspace{-0.1cm}
\begin{description}
\small
\item[Inputs:] $\mathcal{T}$---Graph of an HLTM, $\bar{\cal U}$---Collection of training documents, $M$---Number of new connections for each hidden unit.
\item[Outputs:] Graph $\mathcal{T'}$ of a corresponding SBM.
\end{description}
\small
\begin{algorithmic}[1]
\State $\mathcal{T'} \gets \emptyset,$ $H_L \gets$ graph of the top latent layer in $\mathcal{T}$
\State $V \gets$ observed variables in $\mathcal{T}$
\State $\mathcal{T'}.add\textunderscore graph(H_L), \mathcal{T'}.add\textunderscore units(V)$

\For{variable $Z$ in $H_L$}
	\State $V_Z \gets$ observed variables in subtree rooted at variable $Z$
	\State $\mathcal{T'}.add\textunderscore edges(Z, V_Z),I \gets \emptyset$
	\For{$V'$ in $(V-V_Z)$}
		\State $Z' \gets$ root of the subtree containing $V'$ 
		\State $I_{Z, V'} \gets I(Z, V'|Z',\bar{\cal U})$
		\State $I.add(I_{Z, V'})$
	\EndFor
	\State $I \gets$ sort($I$, `descend')
	\For{$V'$ in the top $M_Z$ pairs in $I$}
	\State $\mathcal{T'}.add\textunderscore edge(V', Z)$
	\EndFor
\EndFor
\State \Return $\mathcal{T'}$
\end{algorithmic}
\caption{{SBM-SFC}($\mathcal{T}$)}\label{alg.SBM-SFC}
\end{algorithm}

\section{Experiments}
In this section we test the performance of our Sparse Boltzmann Machines on three text datasets of different scales: NIPS proceeding papers\footnote{Available at http://www.cs.nyu.edu/~roweis/data.html}, CiteULike articles\footnote{Available at http://www.wanghao.in/data/ctrsr\textunderscore datasets.rar}, and New York Times dataset\footnote{Available at http://archive.ics.uci.edu/ml/datasets/Bag+of+Words}. Experimental results show that SBMs perform consistently well over the three datasets in terms of model generalizability, and SBMs always give much better interpretability. 
\vspace{-0.2cm}

\begin{table*}
\caption{Average per-word perplexity achieved by different methods on different datasets.}
\begin{center}
\begin{tabular}{l cc cc cc}
\toprule
 & \multicolumn{2}{c}{\textbf{NIPS}} &
\multicolumn{2}{c}{\textbf{CiteULike}} & \multicolumn{2}{c}{\textbf{New York Times}}\\\hline
 &Validation  & Test  &  Validation& Test  & Validation & Test\\\hline
\myrowcolour
RS$^*$			& 518 & 547 & 591 & 636 & 1,865 & 1,809 \\ \hline

RS$^+$ 			& 505 & 538 & 795 & 913 & 2,129 & 1,985 \\\hline
\myrowcolour
RS$^{+}$ SFC 	& 532 & 551 & 632 & 668 & 2,021 & 1,910 \\\hline

RS$^+$ Pruned 	& 542 & 565 & \textbf{534} & \textbf{584} & 1,697 & 1,608 \\\hline
\myrowcolour

SBM-SFC		& \textbf{476} & \textbf{488} & 545 & 597 & \textbf{1,624} & \textbf{1,583}\\
\bottomrule
\end{tabular}
\label{table.scores}
\end{center}
\vspace{-0.6cm}
\end{table*}

\subsection{Datasets}
\vspace{-0.1cm}
NIPS proceeding papers consist of 1,740 NIPS papers published from 1987 to 1999. We randomly sample 1,640 papers as training data, 50 as validation data and the remaining 50 as test data. We pre-process the data and choose 1,000 most frequent words throughout the corpus. In this way each document is represented as a vector of 1,000 dimensions, with each element being the number of times the word appears in current document.

CiteULike article collection contains 16,980 articles. Similarly, we randomly divide it into training data with 12,000 articles, validation data with 1,000 articles and test data with 3,980 articles. 2,000 words with highest average TF-IDF values are chosen to represent the articles.

The New York Times dataset includes 300,000 documents, among which we randomly pick 290,000 documents for training, 1,000 for validation and 9,000 for testing. 10,000 words with highest average TF-IDF values are chosen to represent the documents.

\subsection{Training}
\vspace{-0.1cm}
We divide the training data into mini-batches for training. The batch sizes of dataset NIPS, CiteULike and New York Times are 10, 100 and 1,000 respectively. Model parameters are updated after each mini-batch. Assuming that going through all the mini-batches counts as one epoch, we set the maximum number of training epochs to 50. And we train all the models using the Contrastive Divergence algorithm with $T=10$ full Gibbs steps.

As for RBM-based Replicated Softmax, we determine the optimal number of hidden units over the validation data with 10 units as the step size. While for Sparse Boltzmann Machines, we firstly train a two-layer HLTM and then increase the number of connections such that every hidden unit is connected to 20\% of the visible units that are most correlated. A mask matrix is applied to the connection matrix after each parameter update so as to force the sparse connectivity. The numbers of hidden units automatically determined by our algorithm are 112, 194 and 326 for dataset NIPS, CiteULike and New York Times respectively.
\vspace{-0.2cm}

\subsection{Evaluations}
\vspace{-0.1cm}
The log-probability on held-out data is used to gauge the generalization performance of Replicated Softmax and Sparse Boltzmann Machines. As exactly computing these value is intractable (due to the partition function), {\it Annealed Importance Sampling (AIS)} \citep{Neal:2001:AIS:599243.599401,salakhutdinov2008quantitative} was used in \cite{NIPS2009_3856} to estimate the partition function of Replicated Softmax. We extend AIS to Sparse Boltzmann Machines in our experiments. In AIS, we use 500 ``inverse temperatures'' $\beta_k$ spaced uniformly from 0 to 0.5, 3,000 $\beta_k$ spaced uniformly from 0.5 to 0.9, and 6,500 $\beta_k$ spaced uniformly from 0.9 to 1.0, with a total of 10,000 intermediate distributions. The estimates are averaged over 100 AIS runs for each held-out document. Then we calculate the average per-word perplexity as $exp(-\frac{1}{N}\sum_{n=1}^{N}\frac{1}{D_{n}}logP({\cal U}_n))$. A smaller score indicates better generalization performance. Due to the high computation cost, we follow the experiments in \cite{NIPS2009_3856} and randomly sample 50 documents from the validation data to calculate the score. While for test, we use all the 50 test documents in NIPS dataset, and randomly sample 500 documents from test data in CiteULike and New York Times datasets.
\vspace{-0.2cm}

\subsection{Results}
\vspace{-0.1cm}
\subsubsection{Overfitting of Fully-Connected RBMs}
\begin{figure*}[t]
\begin{center}
\begin{tabular}{c}
\mbox{
        \epsfxsize=5cm
        \epsffile{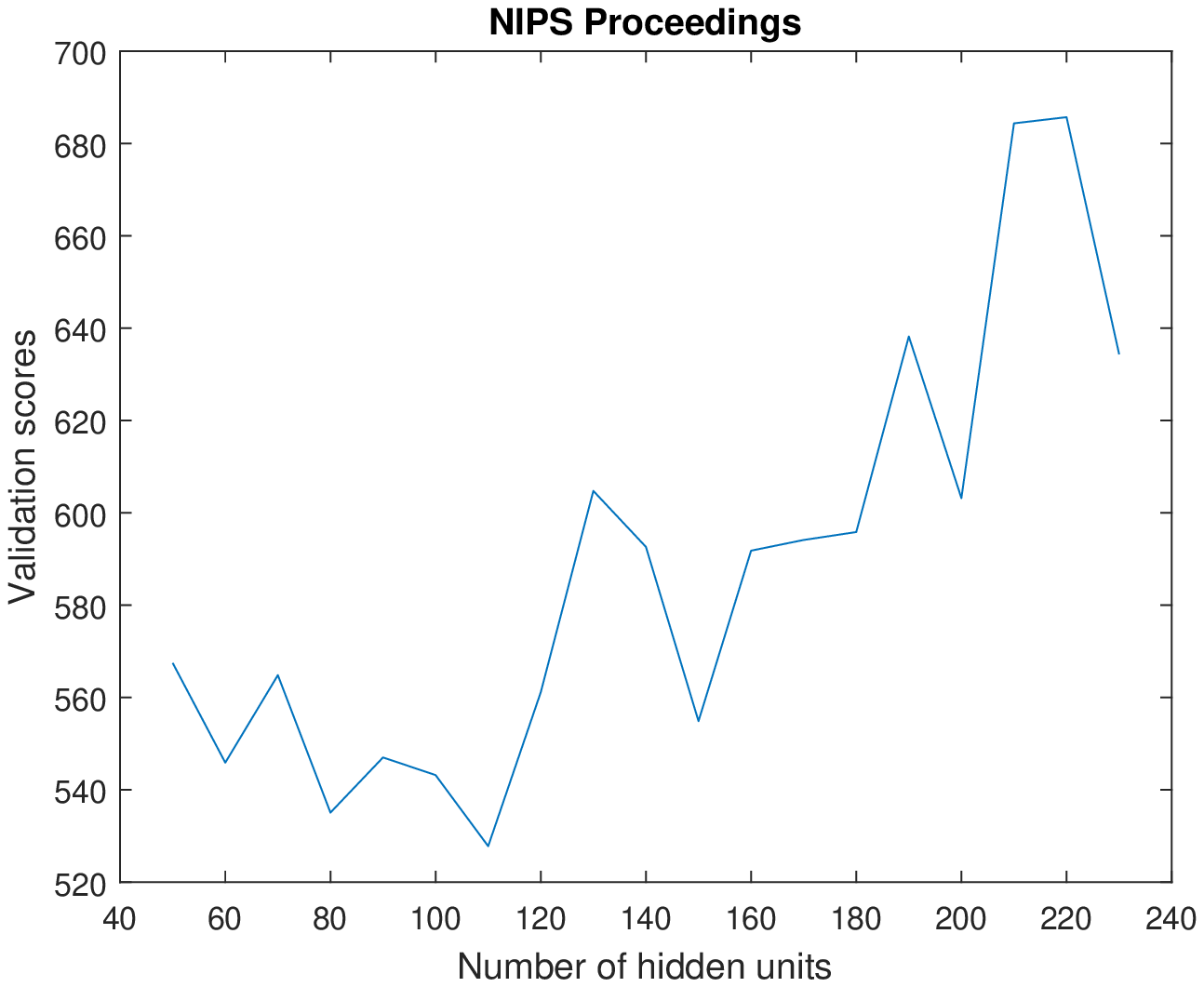}

        \epsfxsize=5cm
        \epsffile{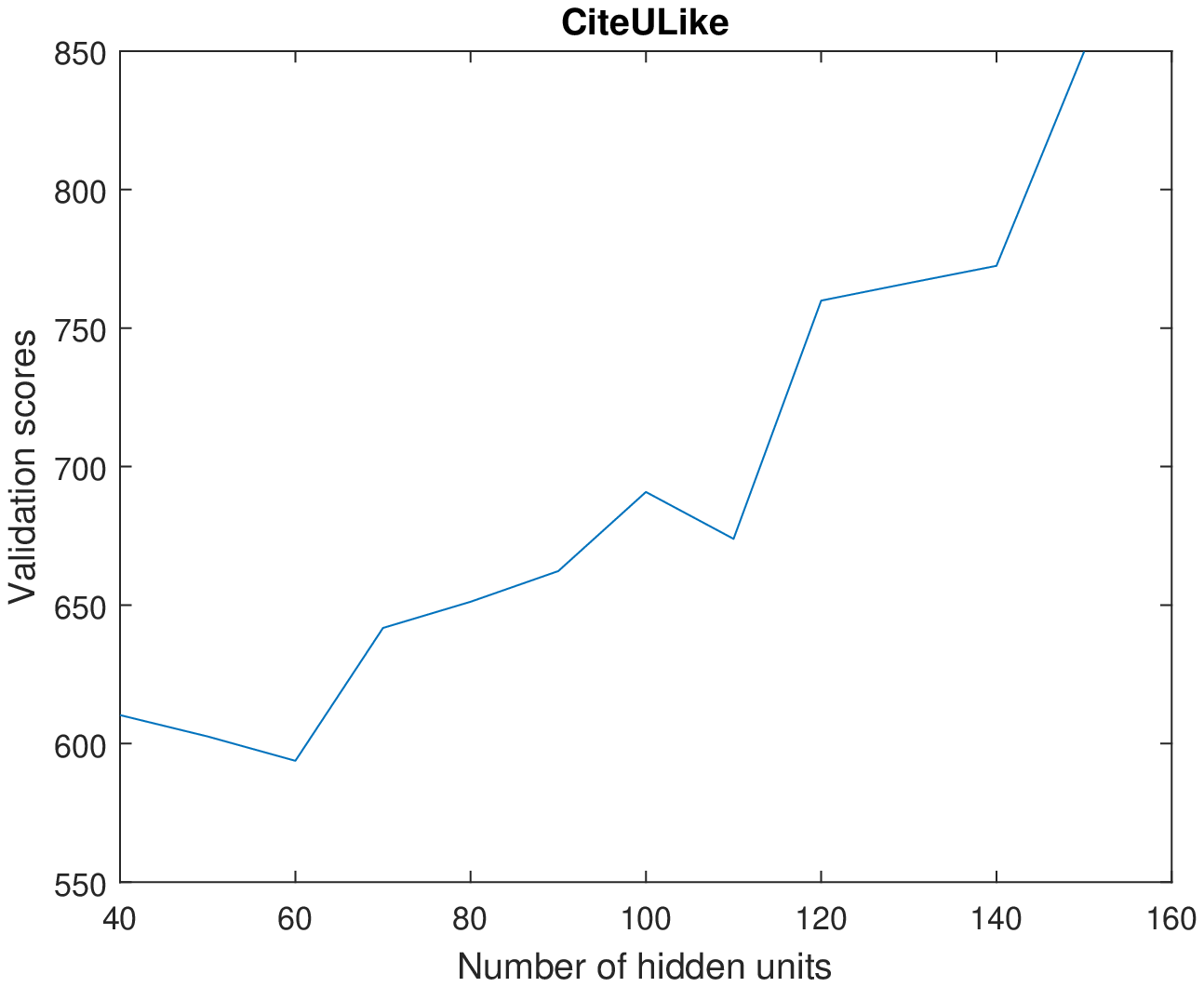}

        \epsfxsize=5cm
        \epsffile{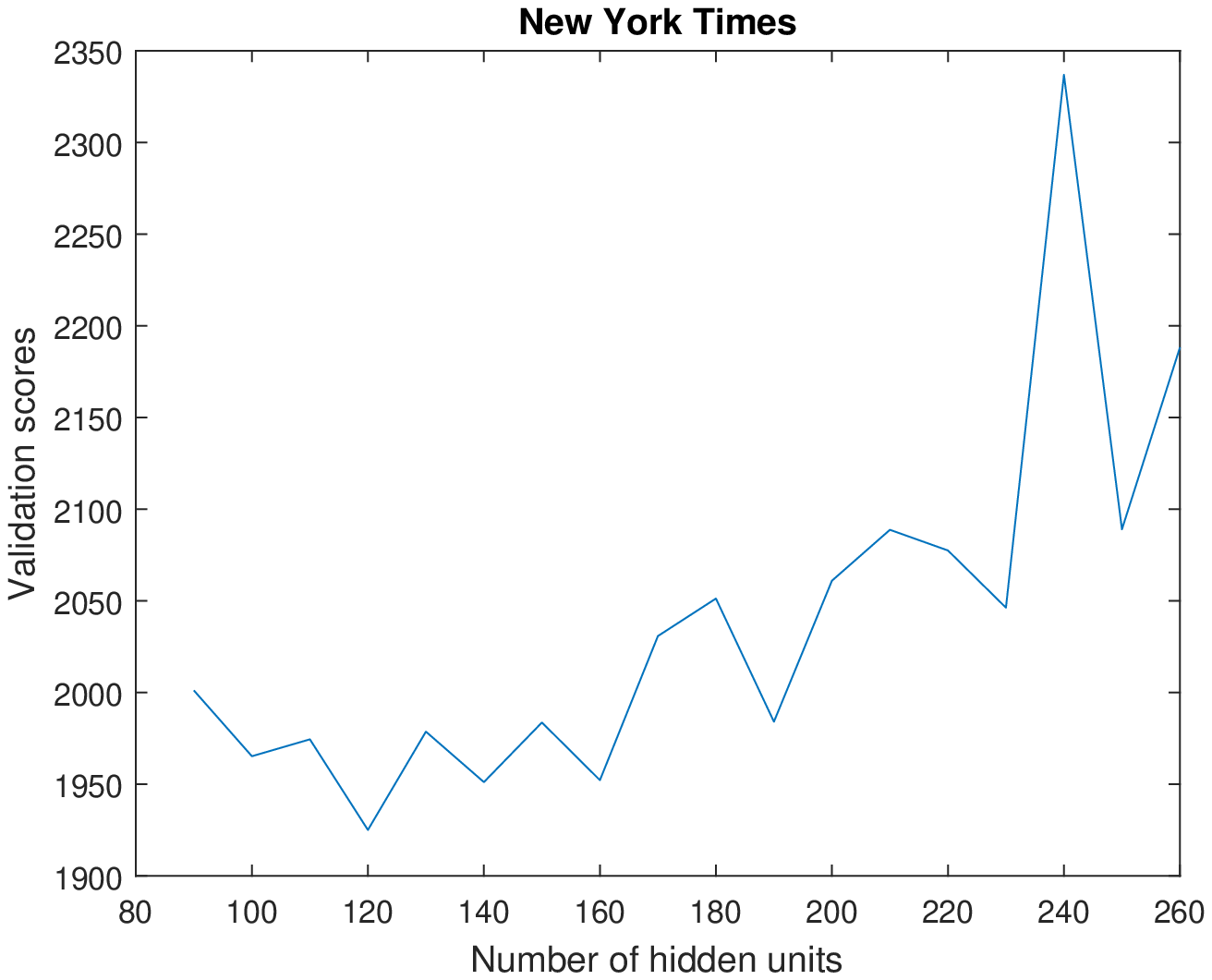}
}
\end{tabular}
\vspace{-0.3cm}
\caption{The generalization performance of Replicated Softmax with different number of hidden units.}
\vspace{-0.5cm}
\label{fig.fc}
\end{center}
\end{figure*}

We first empirically show that, the fully-connected structure in Replicated Softmax can easily lead to overfitting once the number of hidden units (and hence the number of parameters) gets too large. Figure \ref{fig.fc} depicts the average perplexity scores over validation data for Replicated Softmax with different number of hidden units after 30 epochs. We can see that, the optimal numbers of hidden units for the three datasets are 110, 60 and 120 respectively. After that, the performances of the models get worse when the numbers of hidden units gradually increase. Therefore, selecting a proper number of hidden units is crucial to Replicated Softmax since they are very likely to overfit the training data. 

\subsubsection{Generalizability of Sparse Boltzmann Machines and Replicated Softmax}
In this part, we compare the generalization performance of Sparse Boltzmann Machines with Replicated Softmax. We denote our method as {\it SBM-SFC}. Two variants of Replicated Softmax included in comparison are {\it RS$^*$} and {\it RS$^+$}. {\it RS$^*$} trains Replicated Softmax with the optimal number of hidden units. {\it RS$^+$} produces Replicated Softmax with the same number of hidden units as {\it SBM-SFC}. Since this number is normally larger than the optimal number, we denote the method as {\it RS$^+$}. As we can see in Table~\ref{table.scores}, {\it SBM-SFC} consistently outperforms {\it RS$^*$} and {\it RS$^+$} over the three datasets. This confirms that Replicated Softmax with full connectivity is prone to overfitting. It also shows that SBMs can lead to better model fit than fully connected RBMs. This is true even when the number of hidden units in RBMs is optimized through held-out validation. Moreover, the poor performance of {\it RS$^+$} shows that the performance gain of {\it SBM-SFC} cannot be attributed to the larger number of hidden units.

\subsubsection{Comparisons with Redundancy Pruning} 
We also compare our method with the redundancy pruning method which produces Replicated Softmax with sparse connections \citep{NIPS2015_5784}. We denote the method as {\it RS$^+$ Pruned}. It starts from a fully trained model, produced by {\it RS$^+$}, and prunes the connections gradually until the number of connections is reduced to be the same as the model by {\it SBM-SFC}. For each hidden unit, it prunes the set of connections with the smallest absolute weight value. Then it retrains the pruned model for 1 epoch, and conducts pruning again. The pruning and retraining process is repeated until the desired sparsity is reached. In our experiments, the pruning process took 80, 40 and 40 epochs on the three datasets respectively. As shown in Table~\ref{table.scores}, {\it SBM-SFC} achieves comparable model fit as {\it RS$^+$ Pruned}. It shows that our structure learning algorithm is effective and can ease the overfitting problem of fully connected structure as well as the pruning method does. Our method has three advantages over {\it RS$^+$ Pruned}. First, the iterative pruning process of {\it RS$^+$ Pruned} is computationally expensive. Second, it does not offer a way to determine the number of hidden units. One can do this using held-out validation, but that would be computationally prohibitive.  Third, as will be seen later, the models produced by {\it RS$^+$ Pruned} are not as interpretable as those obtained by our method.

\subsubsection{Necessity of Hidden Connections}
In SBMs, we impose a tree structure among the hidden units. Is this necessary?  To answer the question, we compare {\it SBM-SFC} with a method for Replicated Softmax denoted as {\it RS$^{+}$ SFC}.  The model produced by {\it RS$^{+}$ SFC} is the same as that by {\it SBM-SFC}, except that there are no connections among the hidden units. As we can see in Table~\ref{table.scores}, {\it SBM-SFC} always performs better than {\it RS$^{+}$ SFC}. This supports our conjecture that the hidden connections are necessary in our models. The result is not surprising. In a multiple layer model, units at a layer are connected via units at higher layers. In a two layer model, there are no higher layers. Hence it is natural to connect the second-layer units directly. To generalize our work to multiple layers, we will need to add connections only among the hidden units at the top layer.

\begin{table}
\caption{Interpretability scores of models: The three models included have the same number of hidden units.}
\begin{center}
\begin{tabular}{l cc cc cc}
\toprule
 & \multicolumn{1}{c}{\textbf{NIPS}} &
\multicolumn{1}{c}{\textbf{CiteULike}} & \multicolumn{1}{c}{\textbf{New York Times}}\\\hline
\myrowcolour
RS$^+$					& 0.1102 & 0.1499 & 0.1407  \\ \hline
RS$^+$ Pruned			& 0.1006 & 0.1449 & 0.1420  \\ \hline
\myrowcolour
SBM-SFC				& \textbf{0.1235} & \textbf{0.1725} & \textbf{0.1433}   \\
\bottomrule
\vspace{-1cm}
\end{tabular}
\label{table.similarity}
\end{center}
\end{table}

\subsubsection{Interpretability of Sparse Boltzmann Machines and Replicated Softmax}
Next we compare the interpretability of Sparse Boltzmann Machines and Replicated Softmax. Here is how we interpret hidden units. For each hidden unit, we sort the words in descending order of the absolute value of the connection weights between the words and the hidden unit. The top 10 words with the highest absolute weights are chosen to characterize the hidden unit. We propose to measure the ``interpretability'' of a hidden unit by considering how similar pairs of words in the top-10 list are. The similarity between two words is determined using a word2vec model \citep{mikolov2013efficient,DBLP:conf/nips/MikolovSCCD13} trained on part of the Google News datasets \footnote{https://code.google.com/archive/p/word2vec/}, where each word is mapped to a high dimensional vector. The similarity between two words is defined as the consine similarity of the two corresponding vectors. High similarity suggests that the two words appear in similar contexts. Let ${\cal L}$ be the list of words representing a hidden unit. We define the {\em compactness} of ${\cal L}$ to be the average similarity between pairs of words in ${\cal L}$.  We also call it the {\em interpretability score} of the hidden unit.   Note that some of the words in ${\cal L}$ might not be in the vocabulary of the word2vec model we use. This happens infrequently. When it does, the words are simply skipped.
\begin{table}[t]
\centering
\caption{Characterizations of selected hidden units in models produced by {\it SBM-SFC}. Only top 5 words are listed.}
{
 \begin{small}
\begin{tabular}{p{1.2cm}|l}
\toprule
\hline
\multirow{3}{*}{NIPS} 
 & spike neuron pruning weight rules \\
 & pixel pca image pixels images  \\
 & markov likelihood conditional posterior probabilities \\\hline
\multirow{3}{*}{CiteULike }
 & models model modeling causal modelling   \\
 & ancestral species selection duplication evolution  \\
 & network networks connected topology connectivity  \\\hline
\multirow{3}{*}{NYtimes}
 & china beijing south\_africa mexican chinese  \\
 & george\_bush laura\_bush bill\_clinton tournament jew  \\
 & gene patient doctor medical physician  \\\hline
\bottomrule
\end{tabular}
\vspace{-0.5cm}
\end{small}
}
\label{table:topics}
\end{table}
Suppose there are $F$ hidden units in a model. Let $C_1, ... C_F$ be the interpretability scores of hidden units.  We define the {\em interpretability score} of the model as: $Q = \frac{1}{F}\sum_{f=1}^{F}C_f$. Obviously the score depends heavily on the number of hidden units.

Table \ref{table.similarity} reports the interpretability scores of the models produced by {\it RS$^+$}, {\it RS$^+$ Pruned} and {\it SBM-SFC}. The models all have the same number of hidden units and hence their interpertability scores are comparable. {\it SBM-SFC} consistently performs the best over the three datasets, showing superior coherency and compactness in the characterizations of the hidden units and thus better model interpretability. Table \ref{table:topics} shows the characterizations of selected hidden units in the models produced by {\it SBM-SFC}. They are clearly meaningful.

\section{Conclusions}
Overfitting in deep models is caused not only by excessive amount of hidden units, but also excessive amount of connections. In this paper we have developed, for models with a single hidden layer, a method to determine the number of hidden units and the connections among the units. The models obtained by the method are significantly better, in terms of held-out likelihood, than RBMs where the hidden and observed units are fully connected. This is true even when the number of hidden units in RBMs is optimized by held-out validation. In comparison with redundancy pruning, our method is more efficient and is able to determine the number of hidden units. Moreover, it produces more interpretable models. In the future, we will generalize the structure learning method to models with multiple hidden layers.

\pagebreak
\bibliographystyle{aaai}
\bibliography{paper}

\begin{thebibliography}{}

\bibitem[\protect\citeauthoryear{Blei, Ng, and
  Jordan}{2003}]{Blei:2003:LDA:944919.944937}
Blei, D.~M.; Ng, A.~Y.; and Jordan, M.~I.
\newblock 2003.
\newblock Latent dirichlet allocation.
\newblock {\em Journal of Machine Learning Research} 3:993--1022.

\bibitem[\protect\citeauthoryear{Chen \bgroup et al\mbox.\egroup
  }{2016}]{DBLP:conf/aaai/ChenZPC16}
Chen, P.; Zhang, N.~L.; Poon, L. K.~M.; and Chen, Z.
\newblock 2016.
\newblock Progressive {EM} for latent tree models and hierarchical topic
  detection.
\newblock In {\em Proceedings of the Thirtieth {AAAI} Conference on Artificial
  Intelligence},  1498--1504.

\bibitem[\protect\citeauthoryear{Cun, Denker, and
  Solla}{1990}]{Cun90optimalbrain}
Cun, Y.~L.; Denker, J.~S.; and Solla, S.~A.
\newblock 1990.
\newblock Optimal brain damage.
\newblock In {\em Advances in Neural Information Processing Systems},
  598--605.

\bibitem[\protect\citeauthoryear{Han \bgroup et al\mbox.\egroup
  }{2015}]{NIPS2015_5784}
Han, S.; Pool, J.; Tran, J.; and Dally, W.
\newblock 2015.
\newblock Learning both weights and connections for efficient neural network.
\newblock In {\em Advances in Neural Information Processing Systems 28}. Curran
  Associates, Inc.
\newblock  1135--1143.

\bibitem[\protect\citeauthoryear{Hassibi, Stork, and
  Com}{1993}]{Hassibi93secondorder}
Hassibi, B.; Stork, D.~G.; and Com, S. C.~R.
\newblock 1993.
\newblock Second order derivatives for network pruning: Optimal brain surgeon.
\newblock In {\em Advances in Neural Information Processing Systems 5},
  164--171.

\bibitem[\protect\citeauthoryear{Hinton and
  Salakhutdinov}{2009}]{NIPS2009_3856}
Hinton, G.~E., and Salakhutdinov, R.~R.
\newblock 2009.
\newblock Replicated softmax: an undirected topic model.
\newblock In {\em Advances in Neural Information Processing Systems 22}.
\newblock  1607--1614.

\bibitem[\protect\citeauthoryear{Hinton \bgroup et al\mbox.\egroup
  }{2012}]{hinton2012deepsppech}
Hinton, G.~E.; Deng, L.; Yu, D.; Dahl, G.~E.; Mohamed, A.-r.; Jaitly, N.;
  Senior, A.; Vanhoucke, V.; Nguyen, P.; Sainath, T.~N.; et~al.
\newblock 2012.
\newblock Deep neural networks for acoustic modeling in speech recognition: The
  shared views of four research groups.
\newblock {\em IEEE Signal Processing Magazine} 29(6):82--97.

\bibitem[\protect\citeauthoryear{Hinton}{2002}]{Hinton:02}
Hinton, G.~E.
\newblock 2002.
\newblock Training products of experts by minimizing contrastive divergence.
\newblock {\em Neural Computation} 14(8):1771--1800.

\bibitem[\protect\citeauthoryear{Krizhevsky, Sutskever, and
  Hinton}{2012}]{krizhevsky2012imagenet}
Krizhevsky, A.; Sutskever, I.; and Hinton, G.~E.
\newblock 2012.
\newblock Imagenet classification with deep convolutional neural networks.
\newblock In {\em Advances in Neural Information Processing Systems},
  1097--1105.

\bibitem[\protect\citeauthoryear{Lecun \bgroup et al\mbox.\egroup
  }{1998}]{Lecun98gradient-basedlearning}
Lecun, Y.; Bottou, L.; Bengio, Y.; and Haffner, P.
\newblock 1998.
\newblock Gradient-based learning applied to document recognition.
\newblock In {\em Proceedings of the IEEE},  2278--2324.

\bibitem[\protect\citeauthoryear{Lee, Ekanadham, and Ng}{2008}]{lee2008sparse}
Lee, H.; Ekanadham, C.; and Ng, A.~Y.
\newblock 2008.
\newblock Sparse deep belief net model for visual area v2.
\newblock In {\em Advances in Neural Information Processing Systems},
  873--880.

\bibitem[\protect\citeauthoryear{Liu, Zhang, and
  Chen}{2014}]{DBLP:conf/pkdd/LiuZC14}
Liu, T.; Zhang, N.~L.; and Chen, P.
\newblock 2014.
\newblock Hierarchical latent tree analysis for topic detection.
\newblock In {\em Machine Learning and Knowledge Discovery in Databases 2014},
  256--272.

\bibitem[\protect\citeauthoryear{Mikolov \bgroup et al\mbox.\egroup
  }{2011}]{mikolov2011strategies}
Mikolov, T.; Deoras, A.; Povey, D.; Burget, L.; and {\v{C}}ernock{\`y}, J.
\newblock 2011.
\newblock Strategies for training large scale neural network language models.
\newblock In {\em IEEE Workshop on Automatic Speech Recognition and
  Understanding},  196--201.

\bibitem[\protect\citeauthoryear{Mikolov \bgroup et al\mbox.\egroup
  }{2013a}]{mikolov2013efficient}
Mikolov, T.; Chen, K.; Corrado, G.; and Dean, J.
\newblock 2013a.
\newblock Efficient estimation of word representations in vector space.
\newblock In {\em International Conference on Learning Representations
  Workshops}.

\bibitem[\protect\citeauthoryear{Mikolov \bgroup et al\mbox.\egroup
  }{2013b}]{DBLP:conf/nips/MikolovSCCD13}
Mikolov, T.; Sutskever, I.; Chen, K.; Corrado, G.~S.; and Dean, J.
\newblock 2013b.
\newblock Distributed representations of words and phrases and their
  compositionality.
\newblock In {\em Advances in Neural Information Processing Systems 26},
  3111--3119.

\bibitem[\protect\citeauthoryear{Murphy}{2012}]{38136}
Murphy, K.~P.
\newblock 2012.
\newblock {\em Machine learning: a probabilistic perspective}.

\bibitem[\protect\citeauthoryear{Neal}{2001}]{Neal:2001:AIS:599243.599401}
Neal, R.~M.
\newblock 2001.
\newblock Annealed importance sampling.
\newblock {\em Statistics and Computing} 11(2):125--139.

\bibitem[\protect\citeauthoryear{Salakhutdinov and
  Murray}{2008}]{salakhutdinov2008quantitative}
Salakhutdinov, R., and Murray, I.
\newblock 2008.
\newblock On the quantitative analysis of deep belief networks.
\newblock In {\em Proceedings of the 25th International Conference on Machine
  Learning},  872--879.

\bibitem[\protect\citeauthoryear{Smolensky}{1986}]{Smolensky:1986:IPD:104279.104290}
Smolensky, P.
\newblock 1986.
\newblock Parallel distributed processing: Explorations in the microstructure
  of cognition, vol. 1.
\newblock chapter Information Processing in Dynamical Systems: Foundations of
  Harmony Theory,  194--281.

\bibitem[\protect\citeauthoryear{Srinivas and Babu}{2015}]{Srinivas2015}
Srinivas, S., and Babu, R.~V.
\newblock 2015.
\newblock Data-free parameter pruning for deep neural networks.
\newblock In {\em Proceedings of the British Machine Vision Conference},
  31.1--31.12.

\bibitem[\protect\citeauthoryear{Srivastava \bgroup et al\mbox.\egroup
  }{2014}]{Srivastava:2014:DSW:2627435.2670313}
Srivastava, N.; Hinton, G.~E.; Krizhevsky, A.; Sutskever, I.; and
  Salakhutdinov, R.
\newblock 2014.
\newblock Dropout: A simple way to prevent neural networks from overfitting.
\newblock {\em Journal of Machine Learning Research} 15(1):1929--1958.

\bibitem[\protect\citeauthoryear{Sutskever, Vinyals, and
  Le}{2014}]{sutskever2014sequence}
Sutskever, I.; Vinyals, O.; and Le, Q.~V.
\newblock 2014.
\newblock Sequence to sequence learning with neural networks.
\newblock In {\em Advances in Neural Information Processing Systems},
  3104--3112.

\bibitem[\protect\citeauthoryear{Wan \bgroup et al\mbox.\egroup
  }{2013}]{wan2013regularization}
Wan, L.; Zeiler, M.; Zhang, S.; Cun, Y.~L.; and Fergus, R.
\newblock 2013.
\newblock Regularization of neural networks using dropconnect.
\newblock In {\em Proceedings of the 30th International Conference on Machine
  Learning},  1058--1066.

\end{thebibliography}

\end{document}